\documentclass[conference]{IEEEtran}
\usepackage{times}

\usepackage[numbers]{natbib}
\usepackage{multicol}
\usepackage[bookmarks=true]{hyperref}

\usepackage{amsmath,amsfonts,amssymb}
\usepackage{graphicx}

\newcommand*{\vertbar}{\rule[-1ex]{0.5pt}{2.5ex}}

\newcommand{\di}[1]{\mathrm{d}#1}

\DeclareMathOperator*{\argmin}{\arg\!\min}
\DeclareMathOperator{\vect}{vec}
\DeclareMathOperator{\diag}{diag}

\begin{document}

\title{Shape control of simulated multi-segment continuum robots via Koopman operators with per-segment projection}

% You will get a Paper-ID when submitting a pdf file to the conference system
% \author{Author Names Omitted for Anonymous Review. Paper-ID 629}

%\author{\authorblockN{Michael Shell}
%\authorblockA{School of Electrical and\\Computer Engineering\\
%Georgia Institute of Technology\\
%Atlanta, Georgia 30332--0250\\
%Email: mshell@ece.gatech.edu}
%\and
%\authorblockA{Twentieth Century Fox\\
%\and
%\authorblockN{James Kirk\\ and Montgomery Scott}
%\authorblockA{Starfleet Academy\\
%San Francisco, California 96678-2391\\

% avoiding spaces at the end of the author lines is not a problem with
% conference papers because we don't use \thanks or \IEEEmembership


\author{\authorblockN{
    Eron Ristich\authorrefmark{1},
    Jiahe Wang\authorrefmark{2},
    Lei Zhang\authorrefmark{2}, 
    Sultan Haidar Ali\authorrefmark{2},
    Wanxin Jin\authorrefmark{2},
    Yi Ren\authorrefmark{2}, and
    Jiefeng Sun\authorrefmark{2}\authorrefmark{3}
}\authorblockA{\authorrefmark{1}Department of Electrical Engineering and Computer Science \\ University of Michigan, Ann Arbor, Michigan 48109, USA}
\authorblockA{\authorrefmark{2}School for Engineering of Matter, Transport and Energy \\
Arizona State University, Tempe, Arizona 85281, USA}
\authorblockA{\authorrefmark{3}Corresponding author: \href{mailto:jiefeng.sun@asu.edu}{jiefeng.sun@asu.edu}}}

\maketitle

\begin{abstract}
Soft continuum robots can allow for biocompatible yet compliant motions, such as the ability of octopus arms to swim, crawl, and manipulate objects. However, current state-of-the-art continuum robots can only achieve real-time task-space control (i.e., tip control) but not whole-shape control, mainly due to the high computational cost from its infinite degrees of freedom. 
In this paper, we present a data-driven Koopman operator-based approach for the shape control of simulated multi-segment tendon-driven soft continuum robots with the Kirchoff rod model. Using data collected from these simulated soft robots, we conduct a per-segment projection scheme on the state of the robots allowing for the identification of control-affine Koopman models that are an order of magnitude more accurate than without the projection scheme. Using these learned Koopman models, we use a linear model predictive control (MPC) to control the robots to a collection of target shapes of varying complexity. Our method realizes computationally efficient closed-loop control, and demonstrates the feasibility of real-time shape control for soft robots. We envision this work can pave the way for practical shape control of soft continuum robots.
\end{abstract}

\IEEEpeerreviewmaketitle

\section{Introduction}

% Introduce soft robots and some of the associated issues
Soft continuum robotic arms, due to their continuous nature and compliant material constitution, are capable of producing biocompatible motions that enable non-destructive manipulation of delicate materials and safer interaction with humans than their rigid counterparts \citep{rus_design_2015}. In particular, rigid robots are restricted by finite degrees of freedom and thus have limited applicability in applications that have diverse geometric constraints, such as in minimally invasive surgery \cite{runciman_soft_2019}, or when adapting to diverse environments \cite{shah_soft_2021}. Despite the geometrical advantages of soft robots, the efficient and robust control of the \textit{shape} of soft robotic systems has been limited due to the difficulty in the high cost of dynamics prediction \cite{george_thuruthel_control_2018}.

% Existing methods; physics based methods; emphasize nonlinearity w.r.t. control and computationally expensive
Physics-based models for soft robotic arms leverage nonlinear models that require high computational costs. Continuum mechanics models such as the Cosserat rod model \cite{rucker_statics_2011, till_real-time_2019, renda_dynamic_2014} or the simpler Kirchhoff rod model \cite{bao_kinematics_2019} result in a system of partial differential equations (PDEs) that is computationally expensive to solve. Industrial finite element methods have also been employed for end-effector control \cite{duriez_control_2013}, but the high spatial and temporal resolution needed to properly resolve dynamics is often too computationally expensive for real-time control. Reduced order models are a popular alternative that greatly improves the computational speed of forward dynamics while preserving accuracy \cite{goury_fast_2018, tonkens_soft_2021}. However, they must rely on nonlinear control techniques that are slow to converge, especially for shape control where reference shapes must be high dimensional to sufficiently represent the desired geometries. 
% Nevertheless, the benefit of physics based models is their generalizability to all possible configurations of the robot, and such models naturally allow for the incorporation of external force terms, assuming one can construct models for their magnitude. \textbf{TODO: LOOK INTO EXISTING PHYSICS BASED ONLINE FORCE IDENTIFICATION}

\begin{figure}
    \centering
    \includegraphics[width=1.\linewidth]{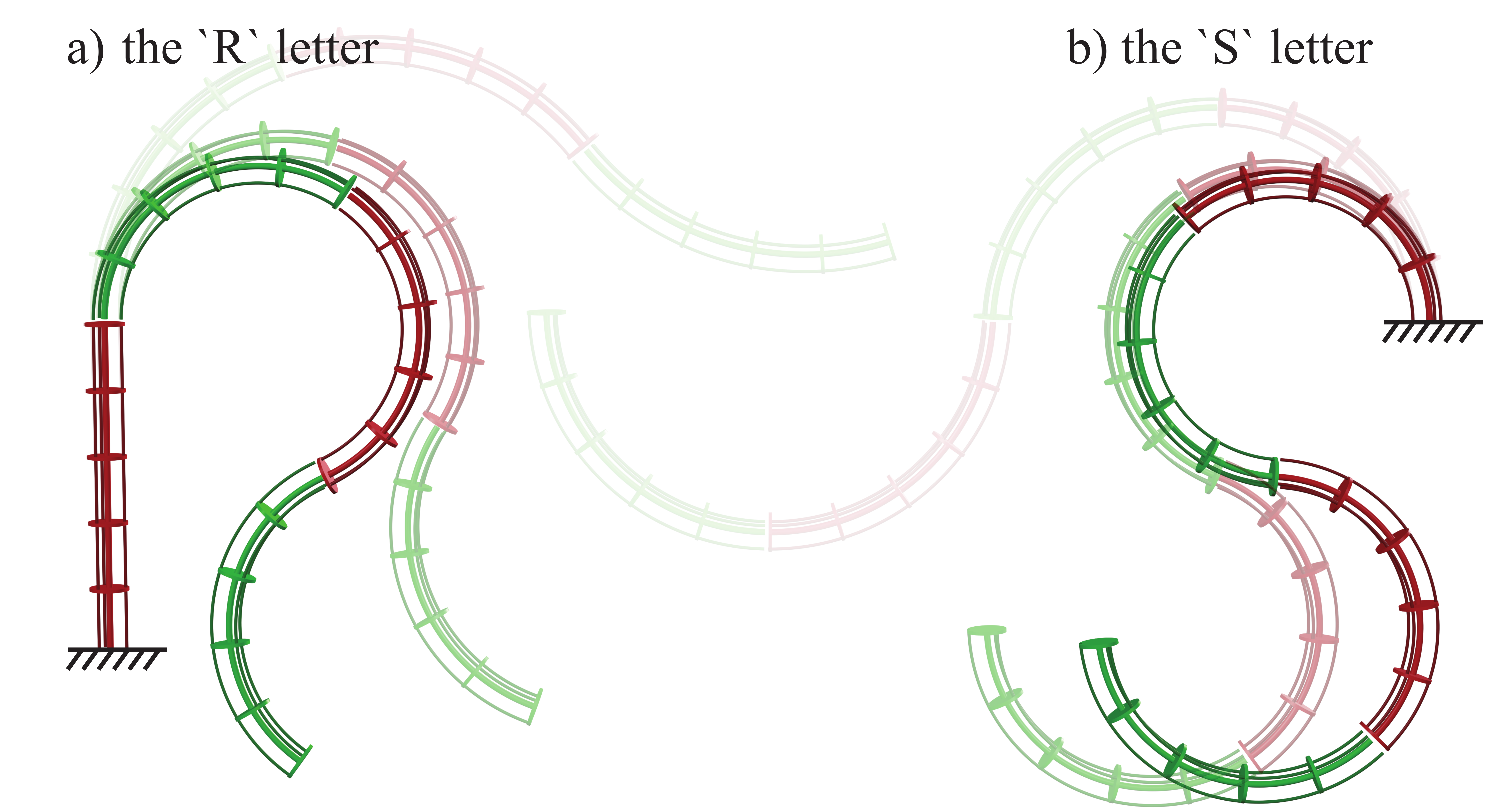}
    \caption{A four-segment tendon-driven soft robot is controlled to move to the shapes of the letters ``R" and ``S".}
    \label{fig:title_fig}
    \vspace{-15pt}
\end{figure}

% Existing methods; data driven methods, variable loading (Bruder); emphasize large data requirements, lack of generalizability outside of training domain
Data-driven models are a promising alternative to physics-based models that learn lightweight learnable operations to approximate the dynamics of a system, resulting in models that are more computationally efficient. Further, as the form of the model can be chosen by a user, they may, by construction, lend themselves to robust control algorithms.
% , and may, by construction, lend themselves to robust control algorithms. 
Deep neural networks have been shown to be an effective choice \cite{thuruthel_model-based_2019, gillespie_learning_2018, zheng_robust_2020}, although resulting controllers are significantly less interpretable than physics-based methods. Further, these heavyweight and nonlinear neural networks are often difficult to integrate into classical control frameworks for fast control and efficient optimization.
% Further, for control tasks, they cannot take advantage of explicit model-based control methods due to their black-box input-output mapping property.
% \textbf{TODO: Look briefly at the paper in the comment following this}
%  Phillip Hyatt, David Wingate, and Marc D Killpack.Model-based control of soft actuators using learned nonlinear discrete-time models. Front. Robot. AI 6: 22. doi: 10.3389/frobt, 2019.
In contrast, \textit{Koopman operator}-based methods construct globally linear models of nonlinear systems, and thus can take advantage of linear control techniques \cite{bruder_advantages_2021, korda_linear_2018}. Recent research has demonstrated the growing success of Koopman-based methods in modeling and controlling soft robotic arms \cite{bruder_modeling_2019, bruder_data-driven_2021, haggerty_control_2023, shi_koopman_2023}. However, these works focus on control of the end effector, and as such, use low-dimensional state variables that are incapable of sufficiently representing the continuous shape of the robotic arm. Ref.~\cite {singh_controlling_2023} successfully leverages Koopman operators to control the shape of soft robotic arms, but makes the strong assumption that arbitrary forces can be applied at several points along the body of the robot. 

% Novel contributions of our work; PI-EDMDc splits into external forces, reduces data requirements of the method, makes it practical to dynamically adjust the external force operator
In this paper, we leverage existing physics-based models to generate trajectory data for a collection of simulated soft robotic arms with realistic actuation forces. We then learn control affine Koopman operator-based models from the projected state variables of these collected trajectories, and then formulate an algorithm for controlling the shape of each soft robotic arm. For example, we can control a four segment robot to move to the shape of letters as in Fig.~\ref{fig:title_fig} (see supplementary video for more examples). The contributions of this work are as follows:
\begin{enumerate}
    \item Development of a per-segment projection scheme for improved accuracy of Koopman models for soft robots. 
    \item Demonstrated shape control of tendon-driven soft robotic arms with between 1 and 5 independent segments.
\end{enumerate}

The rest of the paper is organized as follows. In Sec.~\ref{sec:mathematical_formulation}, we provide relevant background information on Koopman operators and discrete Koopman operator identification. In Sec.~\ref{sec:control_formulation}, we describe additional considerations made for the linear model predictive control (MPC) algorithm used in this work.
% In Sec.~\ref{sec:control_formulation}, we describe a sampling-based control algorithm based on the work of \citet{williams_information-theoretic_2018} which enables light-weight control of the resulting high-dimensional system with low-dimensional control inputs. 
In Sec.~\ref{sec:numerical_experiments}, we describe the simulated experimental setup, discuss the results of the control algorithm as applied to the simulated robots, 
% and point out some notable fail cases
and point out some limitations of the method. Lastly, in Sec.~\ref{sec:conclusion}, we provide some concluding remarks.

\section{Mathematical Formulation} \label{sec:mathematical_formulation}

In this section, we provide a brief overview of applied Koopman operator theory and algorithms for data-driven identification of discrete-time Koopman operators. 

\subsection{Discrete-time Koopman Operators and Identification} \label{sec:disc_time}
Consider a dynamical system whose time evolution is described by
\begin{equation} \label{eq:x_diff_eq}
    \dot{\mathbf{x}} = \mathbf{f}(\mathbf{x}, \mathbf{u}),
\end{equation}
where $\mathbf{x} \in \mathbb{R}^n \equiv \mathcal{X}$ is the state of the system, $\dot{\mathbf{x}}$ denotes its time derivative, $\mathbf{u} \in \mathbb{R}^m \equiv \mathcal{U}$ are the control inputs of the system, and $\mathbf{f}: \mathcal{X} \times \mathcal{U} \rightarrow \mathcal{X}$ is a vector field describing the dynamics. In the case where the dynamics of $\mathbf{x}$ is described by a system of PDEs, then $\mathbf{f}$ can be considered an implicit function of the relevant derivatives of $\mathbf{x}$ and $\mathbf{u}$. In discrete time, the system is propagated forward by the flow map $\mathbf{F}_t: \mathcal{X} \times \mathcal{U} \rightarrow \mathcal{X}$ defined by
\begin{align} \label{eq:x_flow_map}
\begin{split}
    \mathbf{x}(t_0 + t) &= \mathbf{F}_t(\mathbf{x}(t_0), \mathbf{u}(t_0)) \\
    &= \mathbf{x}(t_0) + \int_{t_0}^{t_0 + t} \mathbf{f}(\mathbf{x}(\tau), \mathbf{u}(\tau)) ~\di{\tau}.
\end{split}
\end{align}
When written using indices, the discrete time update can be written succinctly as
\begin{equation}
    \mathbf{x}_{k+1} = \mathbf{F}_t(\mathbf{x}_k, \mathbf{u}_k).
\end{equation}

The discrete-time Koopman operator $\mathcal{K}_t$ is an infinite-dimensional linear operator that acts on functions $g: \mathcal{X} \times \mathcal{U} \rightarrow \mathbb{R}$ called \textit{observables} which belong to some infinite-dimensional Hilbert space $\mathcal{H}$. In particular, the Koopman operator propagates state measurements forward in time using the forward flow map as
\begin{equation} \label{eq:koopman_def}
    \mathcal{K}_t g = g \circ \mathbf{F}_t,
\end{equation}
where $\circ$ is the function composition operator. A common assumption to make Koopman operators amenable to linear control techniques is to assume that control inputs do not evolve forward over a single interval of time, i.e. $\dot{\mathbf{u}} = 0$ over $(t_0, t_0+t)$ \cite{bruder_data-driven_2021, bruder_koopman-based_2021}. In robotics, 
% it is often the case that control inputs are not coupled with the robot dynamics, and as such, 
it is convenient to assume that the control inputs are not coupled with the robot dynamics, i.e, the time evolution of control inputs can be ignored or considered trivial. This is the case especially if they are fully specified by the user. By this assumption, we arrive at
\begin{equation} \label{eq:koopman_observable_update}
    \mathcal{K}_t g (\mathbf{x}_k, \mathbf{u}_k) = g(\mathbf{x}_{k+1}, \mathbf{u}_k).
\end{equation}

Computing the full Koopman operator is infeasible as it is infinite dimensional. However, due to the fact that a Koopman-invariant subspace is spanned by a finite set of eigenfunctions of the Koopman operator, one can construct a globally linear representation of the nonlinear system using a finite set of functions \cite{kaiser_data-driven_2021}. 
% In general, however, these functions may be difficult to identify. 
% In discrete-time, the Koopman eigenfunction $\varphi:\mathbb{R}^n\times\mathbb{R}^m\rightarrow\mathbb{R}$ that corresponds to scalar eigenvalue $\mu$ satisfies the relation
% \begin{equation} \label{eq:discrete-time_koopman_eigenfunction}
%     \varphi(\mathbf{x}_{k+1}, \mathbf{u}_k) = \mathcal{K}_t \varphi(\mathbf{x}_k, \mathbf{u}_k) = \mu \varphi(\mathbf{x}_k, \mathbf{u}_k).
% \end{equation}
To identify operators associated with these eigenfunctions, we first learn approximations of the Koopman operator on a finite set of basis functions. Consider a dataset $D~=~\{\mathbf{x}(t_i), \mathbf{x}(t_i+\Delta t), \mathbf{u}_i\}_{i=1}^N$ and a dictionary of scalar observables $\boldsymbol{\theta}(\mathbf{x}, \mathbf{u}): \mathcal{X} \times \mathcal{U} \rightarrow \mathbb{R}^M$. Dataset $D$ yields three data matrices by stacking their respective column vectors, namely the state $X$, the next state $X^\prime$ after time shift $\Delta t$, and lastly the control input $U$. One can construct data matrices $\Theta(X, U)$ and $\Theta(X^\prime, U)$ where
\begin{equation} \label{eq:theta_matrix}
    \Theta(X, U) = \begin{bmatrix}
        \vertbar & & \vertbar \\
        \boldsymbol{\theta}(\mathbf{x}(t_1), \mathbf{u}_1) & \hdots & \boldsymbol{\theta}(\mathbf{x}(t_N), \mathbf{u}_N) \\
        \vertbar & & \vertbar
    \end{bmatrix},
\end{equation}
and $\Theta(X^\prime, U)$ is similarly defined, except applied to the state after time shift $\mathbf{x}(t_i + \Delta t)$. Then, a finite approximation of the discrete-time Koopman operator, denoted $K_{\Delta t}$, can be found by solving the related least squares problem
\begin{align} \label{eq:discrete-time-koop}
\begin{split}
    K_{\Delta t} & := \argmin_{K_{\Delta t}^*} \left\|K_{\Delta t}^* \Theta(X, U) - \Theta(X^\prime, U) \right\|_2^2 \\
    &\approx \Theta(X^\prime, U) \Theta^\dagger(X, U),
\end{split}
\end{align}
where $(\cdot)^\dagger$ denotes the Moore-Penrose pseudoinverse. In this way, the left-eigenvectors of $K_{\Delta t}$ will be related to the finite set of eigenfunctions of the Koopman operator that span a Koopman-invariant subspace.
% To construct corresponding Koopman eigenfunctions, we can approximate them as linear combinations of the dictionary functions in $\theta$ given by the left-eigenvectors of $K_{\Delta t}$. In particular, for some left-eigenvector $\boldsymbol{\xi} \in \mathbb{R}^M$ with corresponding eigenvalue $\lambda$, the corresponding eigenfunction $\varphi(\mathbf{x}, \mathbf{u})$ can be approximated as 
% \begin{equation}
%     \varphi(\mathbf{x}, \mathbf{u}) = \boldsymbol{\xi}^\intercal \boldsymbol{\theta}(\mathbf{x}, \mathbf{u}).
% \end{equation}
This method is known as extended dynamic mode decomposition (EDMD) \cite{williams_datadriven_2015}. 
% TODO switch some \mathcal{K}_t to just K to denote finite approximation
% Although one might choose to propagate only the learned eigenfunctions associated with $K_{\Delta t}$, explicitly choosing which eigenfunctions to propagate forwards in time delegates responsibility of selecting the most important physical modes of the system to the user. In general, which modes to select is highly problem dependent and difficult to select. For simplicity, 
Eigenfunctions and eigenmodes can be directly extracted from $K_{\Delta t}$ for more control over the model, but for simplicity, 
as done by \cite{bruder_modeling_2019, bruder_data-driven_2021}, we directly use $K_{\Delta t}$ to propagate the dictionary of functions $\boldsymbol{\theta}$ forwards in time, as described by
% can choose to either propagate the identified eigenfunctions forwards in time as described by Eqn.~\eqref{eq:discrete-time_koopman_eigenfunction}, or 
\begin{equation} \label{eq:update_rule}
    \boldsymbol{\theta}(\mathbf{x}_{k+1}, \mathbf{u}) \approx K_{\Delta t} \boldsymbol{\theta}(\mathbf{x}_{k}, \mathbf{u}).
\end{equation}

\subsection{Practical Considerations}
Computing the pseudoinverse in Eqn.~\eqref{eq:discrete-time-koop} may be prone to overfitting, and is sensitive to outliers and noisy data during training \cite{seheult_robust_1989}. To mitigate this issue, as suggested by \citet{bruder_modeling_2019}, we use the Least Absolute Shrinkage and Selection Operator (LASSO) \cite{tibshirani_regression_1996}. This approach improves the sparsity of learned matrices by using an $L^1$ regularization, which is capable of driving terms to 0 and also tends to reduce the magnitude of eigenvalues, thereby improving the stability of the learned system. 
We modify Eqn.~\eqref{eq:discrete-time-koop} as
\begin{align}
\begin{split}
    K_{\Delta t} & := \argmin_{K_{\Delta t}^*} \left\|K_{\Delta t}^* \Theta - \Theta^\prime \right\|_2^2  + \alpha \|K_{\Delta t}^*\|_1,
\end{split}
\end{align}
where $\alpha$ is a hyperparameter that controls the magnitude of the regularization term, and we use the shorthand $\Theta \equiv \Theta(X, U)$ and $\Theta^\prime \equiv \Theta(X^\prime, U)$.

\section{Control Formulation} \label{sec:control_formulation}

% Identified Koopman operator models, especially in low data limits, tend to yield spurious eigenvalues \cite{proctor_generalizing_2018}, and as a consequence, tend to have instability issues when used na{\"i}vely. In particular, due to model inaccuracies, the steady states of the data-driven model for a fixed control input may be prohibitively inaccurate in comparison to the true steady states of the system \cite{haggerty_control_2023}. As such, a popular choice for Koopman-operator based control in soft robotics is linear model predictive control (MPC) \cite{bruder_koopman-based_2021}, which yields a quadratic program and features a finite model look ahead, reducing the consequences of steady state error and instability.

Model predictive control (MPC) is a model-based controller design that features a finite time horizon, allowing the controller to anticipate future events. Importantly, for control affine systems, MPC yields a convex quadratic program which can be solved iteratively and with greater computational efficiency than their nonlinear counterpart \cite{korda_linear_2018}. To enable the use of this technique, we enforce our learned Koopman operator model to be control affine, namely by setting the dictionary of observables to be a concatenation of potentially nonlinear functions of $\mathbf{x}$ with the identity function over $\mathbf{u}$. In particular, we let
\begin{equation} \label{eq:theta_true}
    \boldsymbol{\theta}(\mathbf{x}, \mathbf{u}) = \begin{bmatrix}
        \boldsymbol{\theta}_{\mathbf{x}}(\mathbf{x}) \\ \mathbf{u}
    \end{bmatrix},
\end{equation}
such that $\boldsymbol{\theta}_{\mathbf{x}}: \mathbb{R}^n \rightarrow \mathbb{R}^{M-m}$ is some vector valued function of $\mathbf{x}$. Under the assumption that control inputs do not evolve forward in time, we may remove the last $m$ rows of the learned operator corresponding to the time evolution of the control inputs, thus guaranteeing a control affine form of the learned Koopman operator, given by
\begin{align} \label{eq:control_affine_form}
\begin{split}
    \boldsymbol{\theta}_{\mathbf{x}}(\mathbf{x}_{k+1}) &= \begin{bmatrix}
        I_{M-m} & 0_{(M-m) \times m}
    \end{bmatrix} K_{\Delta t} \begin{bmatrix}
        \boldsymbol{\theta}_{\mathbf{x}}(\mathbf{x}_k) \\ \mathbf{u}_k
    \end{bmatrix} \\
    &= A \boldsymbol{\theta}_{\mathbf{x}}(\mathbf{x}_k) + B \mathbf{u}_k,
\end{split}
\end{align}
where $A \in \mathbb{R}^{(M-m) \times (M-m)}$, $B \in \mathbb{R}^{(M-m) \times m}$, and 
\begin{equation}
    K_{\Delta t} = \begin{bmatrix}
        A & B \\
        0_{m \times (M-m)} & I_{m}
    \end{bmatrix}.
\end{equation}

Lastly, we define the MPC cost as the summation over a horizon of length $H$ given by
\begin{equation} \label{eq:mpc_error}
    \sum_{k=1}^H (
        C \mathbf{z}_{k+1} - \mathbf{r}_{k+1}
    )^\intercal Q (C \mathbf{z}_{k+1} - \mathbf{r}_{k+1}) +
    \Delta \mathbf{u}_{k}^\intercal R \Delta \mathbf{u}_{k}.
\end{equation}
Here, $Q \in \mathbb{R}^{n \times n}$ and $R \in \mathbb{R}^{m \times m}$ are positive definite weight matrices chosen to penalize the different components of the state and control input vectors. Additionally, $\mathbf{z}_k$ is the $k$th lifted state $\mathbf{z}_k = \boldsymbol{\theta}_{\mathbf{x}}(\mathbf{x}_k)$, $\mathbf{r}_k \in \mathcal{X}$ is a reference shape, $\Delta \mathbf{u}_k=\mathbf{u}_{k+1}-\mathbf{u}_k$ is the change in control input, and $C$ is a projection matrix such that $\mathbf{x}_k = C \mathbf{z}_k$. To guarantee that such $C$ exists, we concatenate the state into the dictionary of functions by letting
\begin{equation}
    C = \begin{bmatrix}
        I_n & 0_{n \times (M - m - n)}
    \end{bmatrix}, \quad \text{and} \quad 
    \boldsymbol{\theta}_{\mathbf{x}}(\mathbf{x}) = \begin{bmatrix}
        \mathbf{x} \\ \boldsymbol{\theta}_{\mathbf{x}}^\prime(\mathbf{x})
    \end{bmatrix},
\end{equation}
for some vector valued function $\boldsymbol{\theta}_{\mathbf{x}}^\prime : \mathbb{R}^n \rightarrow \mathbb{R}^{M - m - n}$. To solve this quadratic program, we use the OSQP algorithm developed by \citet{stellato_osqp_2020}.

\begin{figure}
    \centering
    \includegraphics[width=1\linewidth]{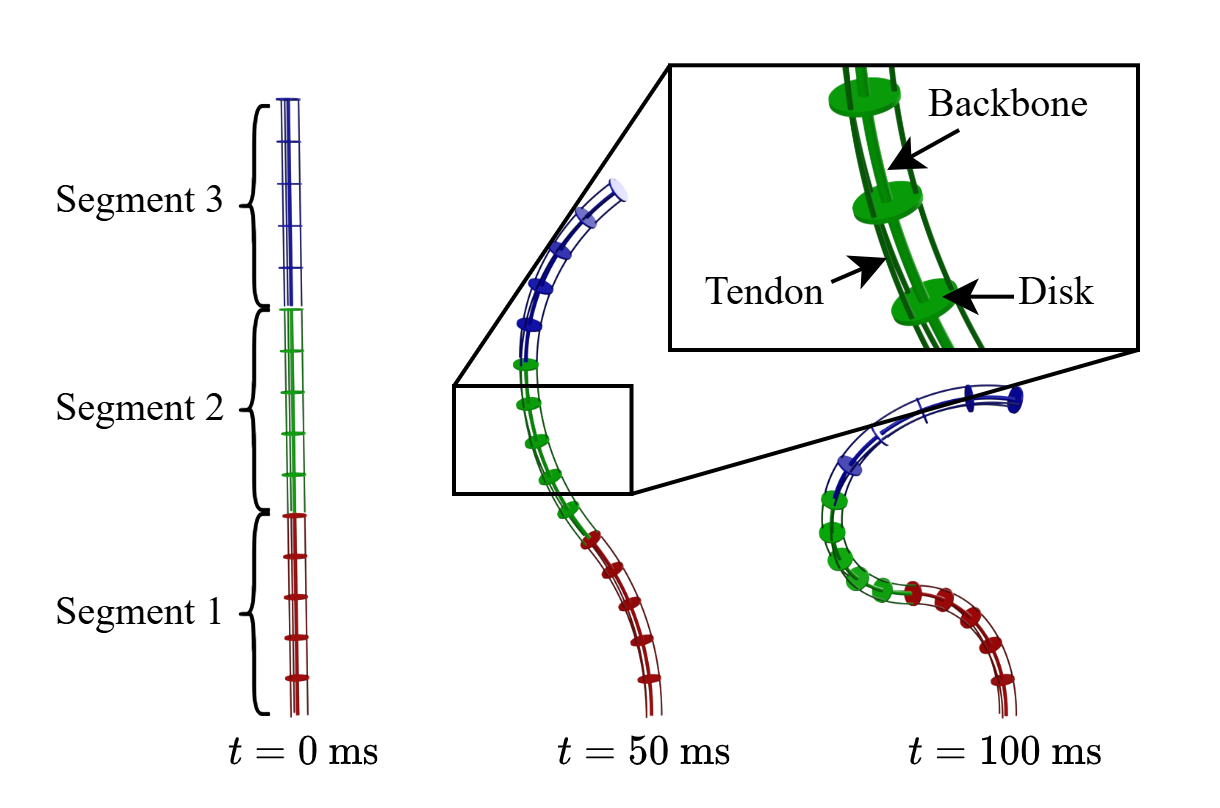}
    \caption{Simulation setup of the 3-segment tendon-actuated robot}
    \label{fig:sim_setup} 
    % \vspace{-15pt} 
\end{figure}

\section{Numerical Experiments} \label{sec:numerical_experiments}

In this section, we conduct and analyze numerical experiments involving the shape control of a simulated tendon-driven soft robotic arm. In particular, we investigate the control of one, two, three, four, and five segment continuum arms, which have increasingly complex possible shapes. 

\subsection{Simulation Setup}

\begin{table}
    \centering
        \caption{Material, geometrical, and simulation properties for each tendon robot configuration}
    \begin{tabular}{|c|c|}
        \hline
        \textbf{Parameter} & \textbf{Value} \\
        \hline
        Length per segment (m) & $1.0$ \\
        Young's modulus (GPa) & 200 \\
        Density (kg/m$^3$) & 8000 \\
        Backbone radius (cm) & $1.0$ \\
        \hline
        Number of tendons per segment & 3 \\
        Tendon offset (cm) & 4 \\
        Tendon angles (deg) & 0, 120, 240 \\
        \hline
        Spatial discretization per segment & 100 \\
        Time step (s) & 0.01 \\
        \hline
    \end{tabular}

    \label{tab:mat_prop} 
   \vspace{-5pt} 
\end{table}

Numerical simulations will be used to collect data for training the Koopman model and used as a testing environment for implementing the control. 
To simulate the soft arm, we separate it into its constituent parts, namely, the backbone, the tendons, and tendon routing disks as shown in Fig.~\ref{fig:sim_setup}. We model the backbone of the soft robot using the Kirchhoff rod model, subject to tendon forces described by \citet{rucker_statics_2011}. In particular, we use the implicit BDF-$\alpha$ method described by \citet{till_real-time_2019} to solve for the continuum robot's dynamics. Under this discretization, the system of PDEs for the tendon actuated robot is 
% \begin{align} \label{eq:kirchhoff_model}
% \begin{split}
%     \mathbf{p}_s &= R \mathbf{e}_3 \\
%     R_s &= R \hat{\mathbf{v}} \\
%     \mathbf{v}_s
%     &= \Phi^{-1} (-\Gamma_u + \Lambda_m) \\
%     \mathbf{q}_s &= -\hat{\mathbf{v}} \mathbf{q} + \hat{\boldsymbol{\omega}} \mathbf{e}_3 \\
%     \boldsymbol{\omega}_s &= \mathbf{v}_t - \hat{\mathbf{v}} \boldsymbol{\omega}
% \end{split}
% \end{align}
\begin{align} \label{eq:kirchhoff_model}
\begin{split}
    \mathbf{p}_t &= R \mathbf{q} \\
    R_t &= R \hat{\boldsymbol{\omega}} \\
    \mathbf{v}_t &= \boldsymbol{\omega}_s + \hat{\mathbf{v}} \boldsymbol{\omega} \\
    \mathbf{q}_t &= \frac{1}{\rho A} R^\intercal \mathbf{f} - \hat{\boldsymbol{\omega}} \mathbf{q} \\
    \boldsymbol{\omega}_t &= (\rho J)^{-1} \left(
        K_{bt} (\mathbf{v}_s - \mathbf{v}_s^*) + \hat{\mathbf{v}} K_{bt} (\mathbf{v} - \mathbf{v}^*) \right.\\
        &\qquad + \left. R^\intercal \mathbf{l} - \hat{\boldsymbol{\omega}} \rho J \boldsymbol{\omega}
    \right)
\end{split}
\end{align}
% subject to the BDF-$\alpha$ semi-discretization given by
% \begin{align} \label{eq:bdf-alpha}
% \begin{split}
%     \mathbf{p}_t &= R \mathbf{q} \\
%     R_t &= R \hat{\boldsymbol{\omega}} \\
%     \mathbf{v}_t &= c_0 \mathbf{v} + \mathbf{v}^h \\
%     \mathbf{q}_t &= c_0 \mathbf{q} + \mathbf{q}^h \\
%     \boldsymbol{\omega}_t &= c_0 \boldsymbol{\omega} + \boldsymbol{\omega}^h
% \end{split}
% \end{align}
where $\mathbf{p}$ is the position of the backbone with respect to arclength $s$, $R$ is the orientation along the backbone, $\mathbf{v}$ is the angular strain, $\mathbf{q}$ is the linear velocity, and $\boldsymbol{\omega}$ is the angular velocity. The subscripts $(\cdot)_s$ and $(\cdot)_t$ denote partial derivatives with respect to arclength $s$ and time $t$. The superscript $(\cdot)^*$ denotes the variables value in the reference configuration. The operator $\hat{\cdot}$ denotes the mapping from $\mathbb{R}^3$ to $\mathfrak{se}(3)$. Further, we let
% \begin{align}
% \begin{split}
%     \Phi &= K_{bt} + c_0 B_{bt} \\
%     \Gamma &= \hat{\mathbf{v}} \mathbf{m}^b + B_{bt} \mathbf{v}_s^h \\
%     \Lambda &= \rho(\hat{\boldsymbol{\omega}} J \boldsymbol{\omega} + J \boldsymbol{\omega}_t) - R^\intercal \boldsymbol{\ell},
% \end{split}
% \end{align}
$\rho$ denote the density of the backbone material, $A$ denote the area of its cross section, $J$ be the matrix of second area moments of the cross section, $K_{bt}$ be the stiffness matrix for bending and torsion, and
$\mathbf{f}, \mathbf{l} \in \mathbb{R}^3$ be the external forces and moments due to external loads and tendon actuation forces.
For more details on the BDF-$\alpha$ method, the evaluation of tendon forces, and all relevant boundary conditions, we refer the reader to \citet{till_real-time_2019}. The solutions to Eqn.~\eqref{eq:kirchhoff_model} are used as the ground truth in the control experiments, and are collected for use as trajectories when learning Koopman operators.
% When learning the continuous-time Koopman operator, one can use the semi-discretized PDE given by the BDF-$\alpha$ method in place of the full dynamical PDE for computational efficiency and consistency with the dynamics of the simulations.

As shown in Fig.~\ref{fig:sim_setup}, the reference shape of the backbone is assumed to be straight and cylindrical, and all tendons are routed parallel to the backbone. Each segment has their own set of tendons which only apply forces along the length of that segment. All other relevant material, geometrical, and simulation properties are summarized in Table ~\ref{tab:mat_prop}.

\subsection{Data Collection}

\begin{figure}
    \centering
    \includegraphics[width=1.0\linewidth]{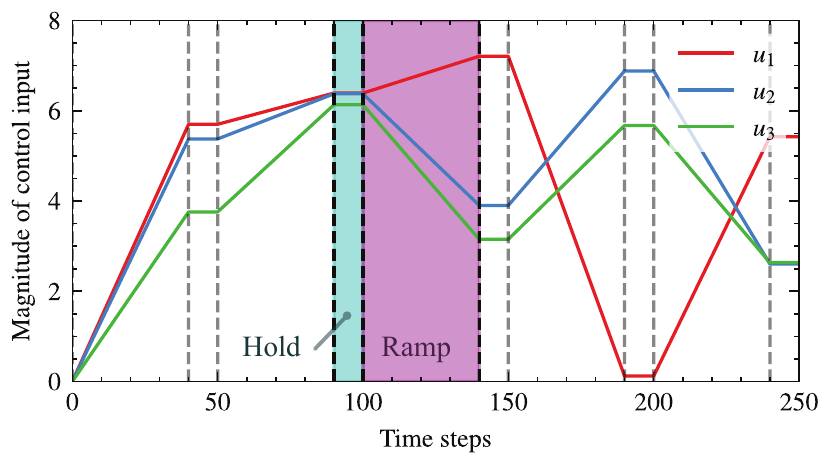}
    \caption{Sample of ramp and hold timescales for control input sequences used when collecting trajectories for training Koopman operators}
    \label{fig:control_inputs} 
    \vspace{-15pt} 
\end{figure}

For each robot, 480 trajectories including 120,000 snapshots $(\mathbf{x}(t), \mathbf{x}(t + \Delta t), \mathbf{u}(t))$ are collected. To collect the data, we first initialize the robot to its steady state in absence of control inputs. In particular, we let the control input $\mathbf{u}$ be a vector of the forces applied to each tendon, and initially set $\mathbf{u}(0) = 0$. Then, we randomly sample 5 unique control inputs from $\mathbb{R}^M$ by sampling each component from a uniform distribution $\mathcal{U}(0, 8)$ in Newtons, where $\mathcal{U}(a, b)$ denotes a uniform distribution between minimum value $a$ and maximum value $b$. We denote these 5 control inputs as $\mathbf{u}_i$ for $i \in \{1, \hdots, 5\}$. Then, as a function of the simulation clock time $t$, we allow control inputs to linearly interpolate or ramp between each consecutive $\mathbf{u}_i$ for time $t_r$ before holding them fixed at $\mathbf{u}_i$ for time $t_h$. Specifically, we define, for $t \in [0, 5 (t_r + t_h)]$,
\begin{equation}
    \mathbf{u}(t) = \begin{cases}
        \mathbf{u}_i (1 - \theta) + \mathbf{u}_{i + 1} \theta & t - t_0 < t_r \\
        \mathbf{u}_{i + 1} & t_r \leq t - t_0 < t_h
    \end{cases},
\end{equation}
where $\tau_0 = 0$, and we define $i = \lfloor t / (t_r + t_h) \rfloor$, $t_0 = (t_r + t_h) i$, and $t_1 = (t_r + t_h) (i + 1)$. The interpolation parameter $\theta$ is defined as
\begin{equation}
    \theta = \frac{t - t_0}{t_1 - t_0}.
\end{equation}

In particular, we define $\Delta t = 0.01$s, $t_r = 40 \Delta t$ and $t_h = 10 \Delta t$. After ramping to and holding at each of the 5 randomly selected control inputs, we then terminate the simulation, and run another. A sample of the control inputs used in one of the 480 trajectories used for the one segment robot is shown in Fig.~\ref{fig:control_inputs}. Snapshots are generated by subsampling from these simulations, where $\mathbf{x}$ is computed as described in the following Sec.~\ref{sec:ne_koop_iden}. 
% This procedure is repeated 480 times.

\subsection{Koopman Identification and Per-segment Projection} \label{sec:ne_koop_iden}

The Koopman operators used are learned using the data-driven identification method described in Sec.~\ref{sec:disc_time}. In line with the control affine assumption given by Eqn.~\eqref{eq:theta_true}, we choose $\boldsymbol{\theta}_{\mathbf{x}}(\mathbf{x})$ to be an impure function yielding two time-delayed coordinates of $\mathbf{x}$, such that
\begin{equation}
    \boldsymbol{\theta}_{\mathbf{x}}(\mathbf{x}_k) = \begin{bmatrix}
        \mathbf{x}_{k} \\ \mathbf{x}_{k-1} \\ \mathbf{x}_{k-2}
    \end{bmatrix}.
\end{equation}

We define the control input vector $\mathbf{u} \in \mathbb{R}^m$ as the tension applied to each tendon, where $m$ is the total number of tendons. The ordering of these control inputs is arbitrary, however, for consistency, they are ordered first by increasing segment index, and then by increasing angle $0^\circ$, $120^\circ$, and $240^\circ$. As a control input to the Koopman operator, one might also choose $\Delta \mathbf{u}_k = \mathbf{u}_{k} - \mathbf{u}_{k-1}$ over $\mathbf{u}_k$ due to the fact that it is zero centered. 

A natural choice of state variable $\mathbf{x} \in \mathbb{R}^n$ is a concatenation of several positions $\mathbf{p}$ along the backbone such that $n = 3
\bar{p}$, where $\bar{p}$ is the number of points subsampled from the backbone. This allows for trivial comparison with a reference shape, as long as the curve is also defined by positions in space. Additionally, position tends to be easier to observe in practice (i.e. with motion capture cameras), and omitting the orientation $R$ and other state variables enables smaller dimensionality of the learned operators.

\begin{figure}
    \centering
    \includegraphics[width=1\linewidth]{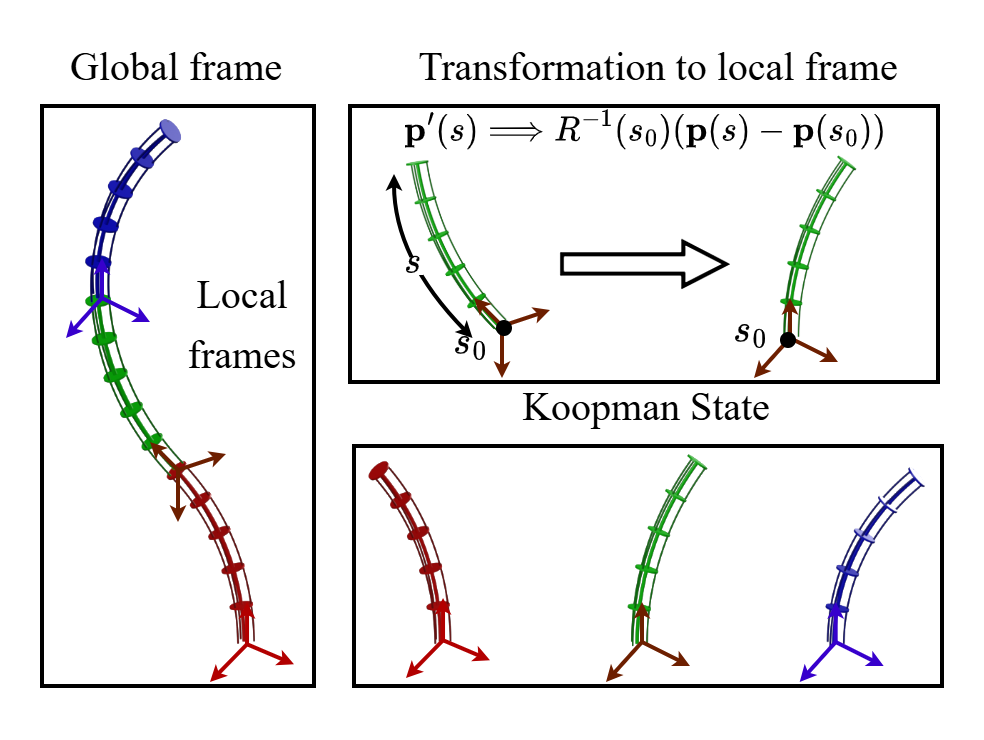}
    \caption{Projection of each segment in the three segment robot onto their local coordinate frame}
    \label{fig:proj_explainer}
    \vspace{-15pt}
\end{figure}

However, na\"ively choosing this set of state variables leads to large model errors due to the control affine assumption. Indeed, small changes in the control input for segments closer to the fixed end of the robotic arm leads to highly nonlinear dynamics at the distal end. Instead, as each segment has their own independent set of tendons, one can project each segment into its local coordinate frame, and use the resulting coordinates to learn the corresponding Koopman operator. This procedure guarantees that the effect of tendon forces applied to a segment closer to the fixed end is minimimal on a segment near the distal end, promoting sparsity of the control input Koopman matrix. In particular, 
% for each segment, let $s_0$ be the position in along the arclength for which the segment originates. Then, we project the segment into the local frame determined by its origin. 
we assign to each position along the arclength $s$ of the backbone,
\begin{equation} \label{eq:projection_explainer}
    \mathbf{p}^\prime(s) \Longrightarrow R^{-1}(s_0) (\mathbf{p}(s) - \mathbf{p}(s_0)),
\end{equation}
where $\mathbf{p}^\prime(s)$ is the transformed position coordinate, and $s_0$ is the position along the arclength of the backbone where the segment originates. This procedure is summarized by Fig.~\ref{fig:proj_explainer}. To further reduce dimensionality, 10 position vectors are taken uniformly from each segment instead of using the full discretization used in the simulations.

\begin{figure}
    \centering
    \includegraphics[width=1\linewidth]{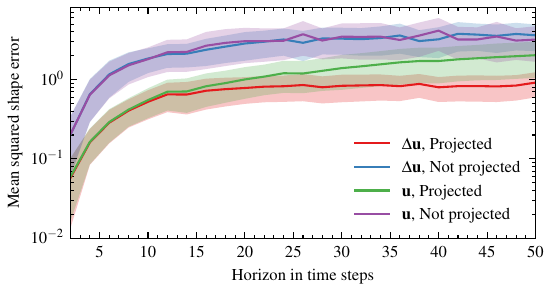}
    \caption{Mean and 1 standard deviation range of the mean squared shape error of Koopman models as a function of the length of horizon. Models are trained in the case of a three segment robot. $\Delta \mathbf{u}$ or $\mathbf{u}$ refers to using the respective value as the control input in the Koopman operator. Projected or not projected refers to using or not using the projection scheme described by Eqn.~\eqref{eq:projection_explainer}}
    \label{fig:koopman-accuracy} 
    \vspace{-15pt} 
\end{figure}

In Fig.~\ref{fig:koopman-accuracy}, we compare the mean squared unprojected shape error of each choice of Koopman model as a function of the prediction horizon. The error is computed over 200 trajectories with 200 time steps, each collected separately from the training of the Koopman operator. We find that using $\Delta \mathbf{u}$ with the aforementioned projection scheme leads to models with the highest accuracy. For consistency, the full state of the non-projected models are reconstructed from the Koopman state by including orientations $R(s_0)$ for each segment and inverting the procedure described by Eqn.~\eqref{eq:projection_explainer}. Error is then computed at each discretized point $s_i$ along the backbone using the mean squared shape error (MSE) given by
\begin{equation}
    \text{MSE} = \frac{1}{n} \sum_{i=1}^n \|\mathbf{p}(s_i) - \mathbf{p}_{\text{ref}}(s_i)\|^2_2,
\end{equation}
where $\mathbf{p}(s)$ denotes the position of the simulated backbone as a function of arclength, and $\mathbf{p}_{\text{ref}}(s)$ is the position of the reference shape as a function of arclength. In all figures the MSE will be plotted on a semilog plot.

\subsection{Results}

In our numerical experiments, we provide a static reference shape and observe the resulting dynamical trajectories the soft robot takes to obtain the correct result. To generate these reference shapes, we solve the corresponding static equilibrium equations derived from Eqn.~\eqref{eq:kirchhoff_model} for some fixed and randomly selected value of $\mathbf{u}$. For each method, we observe and compare the time to convergence and the quality of the trajectory in the neighborhood of the reference shape.

\begin{figure*}
    \centering
    \includegraphics[width=1\linewidth]{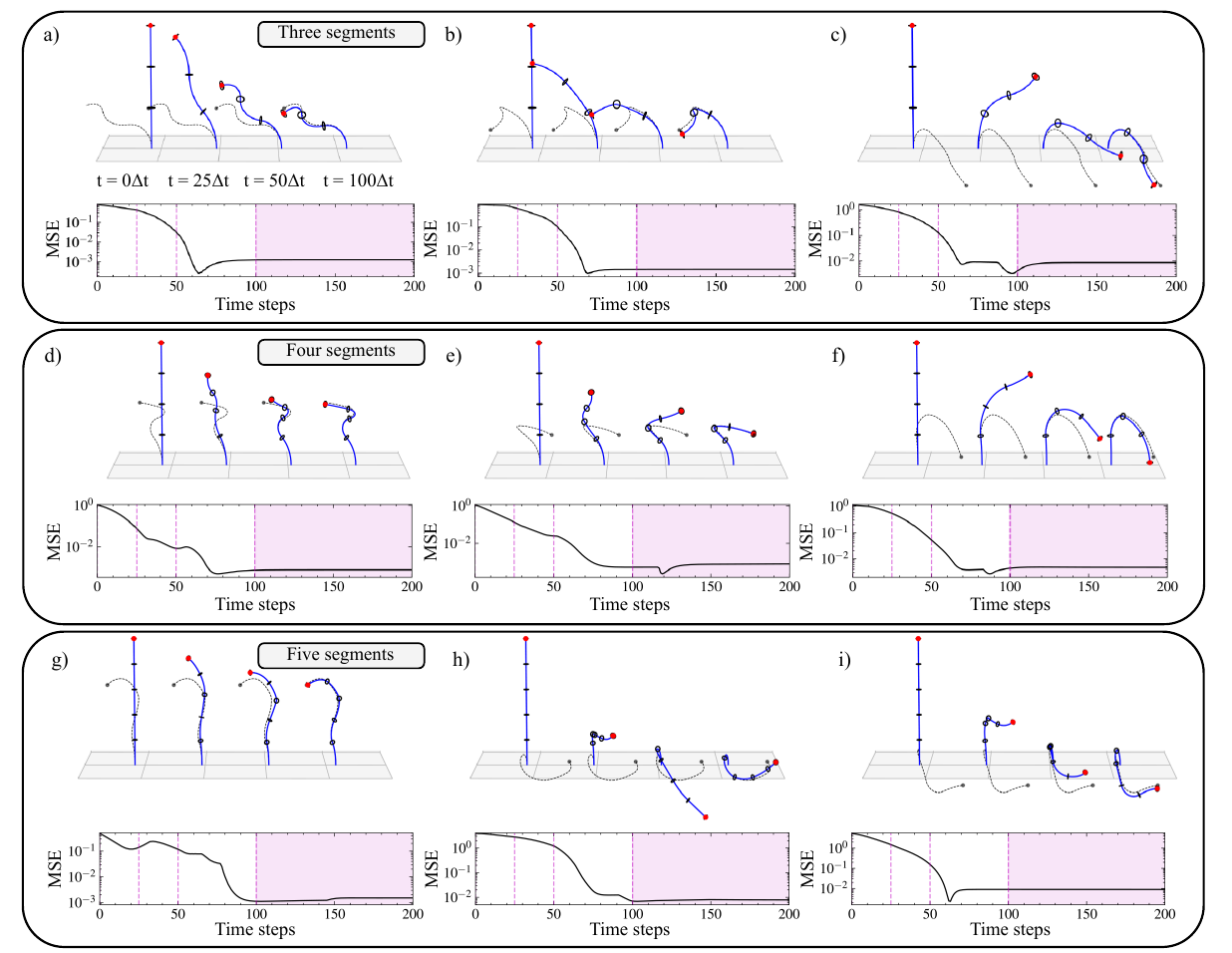}
    \caption{Sample trajectories of the 3 (a-c), 4 (d-f), and 5 (g-i) segment robots shown in blue for three different reference shapes shown as dashed black lines (see supplementary video). Plotted below is the mean squared shape error as a function of the number of time steps. Dashed pink lines denote where the trajectory snapshots are taken. The highlighted pink regions denote the 2nd half of the simulation, where most trajectories have already converged. }
    \label{fig:s5results}
\end{figure*}

% Figure 1: example shape control of 5 segments
Fig.~\ref{fig:s5results} showcases several snapshots of the backbone and the loss curve for three example trajectories from the 3, 4, and 5 segment robots. In each case, the final shape of the robot is qualitatively and quantitatively very close to the reference shape, demonstrating the efficacy of the developed method. The shaded pink regions highlight the 2nd half of the simulation, which showcases the different types of observed convergence behavior. In Fig.~\ref{fig:s5results}.g, a jump in the error is observed at around 150 time steps, corresponding to large dynamical errors in the Koopman model. In several cases, such as in Fig.~\ref{fig:s5results}.a, b, c, and i, the Koopman model error causes a spike in the shape error between 60 and 70 time steps. In Fig.~\ref{fig:s5results}.e we see this spike occurs in the 2nd half of the simulation after the robot appears to have initially converged. Despite these model errors affecting the stability, the final shape error is consistently improved by over 2 orders of magnitude. 

% Figure 2: convergence pattern
\begin{figure*}
    \centering
    \includegraphics[width=1\linewidth]{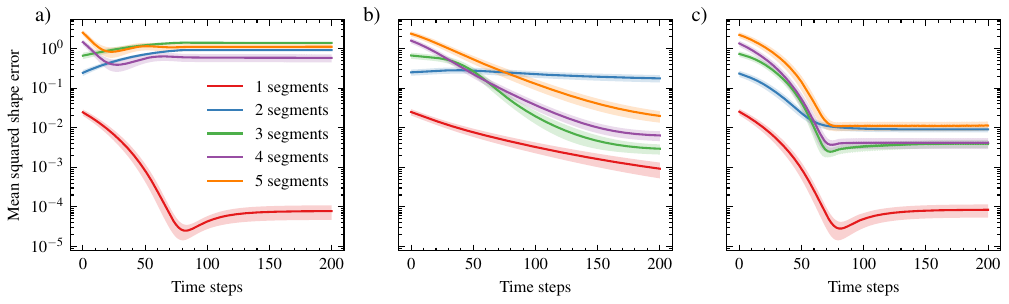}
    \caption{Mean and 0.25 standard deviation range of the mean squared shape error for different methods. a) is the shape error using a weighted $Q$ matrix described by Eqn.~\eqref{eq:opt_q_matrix} without using the projection scheme. b) uses $Q$ proportional to the identity matrix with the projection scheme, while c) uses a weighted $Q$ matrix described by Eqn.~\eqref{eq:opt_q_matrix} with the projection scheme. Quarter standard deviations are chosen to ensure nonnegativity on the semi-log plot.}
    \label{fig:convergence}
\end{figure*}

For each robot and for each of three tested controllers, 200 simulations of 200 time steps each are conducted and the shape error of each controller is computed. First, we test the controller with a model that does not use the projection scheme. Then, we test two controllers using the projection scheme that use two different choices for the gain matrix $Q$. As shown in Fig.~\ref{fig:convergence}, the control algorithm with the projection method is able to consistently improve the shape error between the robot and a reference shape by several orders of magnitude over a period two seconds in real time. When the projection method is not used, the shape error of the multiple segment robots is not able to improve as seen in Fig.~\ref{fig:convergence}.a. Further, when using the projection method, the convergence behavior is heavily dependent on the choice of gain matrix $Q$. Fig.~\ref{fig:convergence}.b and Fig.~\ref{fig:convergence}.c showcase two principled choices of $Q$, where $Q$ is a rescaled identity matrix $Q = 20 I$ and where $Q$ is a weighted diagonal matrix respectively. The weighted $Q$ matrix features larger terms for segments closer to the fixed end, and smaller terms for segments near the distal end, given by 
% Eqn.~\eqref{eq:opt_q_matrix}
\begin{equation} \label{eq:opt_q_matrix}
    Q = 20 (\diag \circ \vect) \left(\begin{bmatrix}
        \bar{p} & \bar{p}-1 & & 1 \\
        \bar{p} & \bar{p}-1 & \hdots & 1 \\
        \bar{p} & \bar{p}-1 & & 1
    \end{bmatrix}\right).
\end{equation}
Here, $\diag$ denotes the diagonalization operation, $\vect$ denotes the vectorization operation. This second choice of $Q$, although promoting faster convergence, demonstrates the steady state model error inherent to Koopman models of this type, described more clearly in Sec.~\ref{sec:limitations}.

\subsection{Limitations} \label{sec:limitations}

\begin{figure}
    \centering
    \includegraphics[width=1\linewidth]{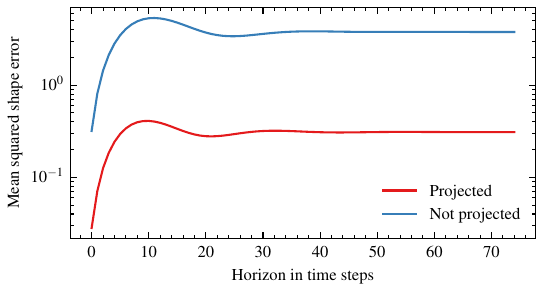}
    \caption{Norm of the steady state error for the Koopman model of the three segment robot when control inputs are fixed at 0 as a function of the number of time steps.}
    \label{fig:steady_err} 
    \vspace{-15pt} 
\end{figure}

Identified Koopman operator models, especially in low data limits, tend to yield spurious eigenvalues \cite{proctor_generalizing_2018}, and as a consequence, tend to have instability issues when used na{\"i}vely. In particular, due to model inaccuracies, the steady states of the data-driven model for a fixed control input may be prohibitively inaccurate in comparison to the true steady states of the system \cite{haggerty_control_2023}. In general, the learned Koopman models, for a fixed control input, are generally unable to converge to the correct steady state, even in trivial cases. For example, in the case where the robot starts in its initial configuration, and no work is done by the control inputs, a significant amount of deviation is observed from the initial configuration, as shown in Fig.~\ref{fig:steady_err}. This effect is reflected by nearly any configuration in the dynamical manifold. As a consequence of these model errors, the MPC controller is unable to maintain the lowest obtainable error as shown in Fig.~\ref{fig:convergence}, which has a large dip in the shape error between 50 and 100 time steps. Principled ways to address this steady state error have begun to appear in literature, such as the work done by \citet{haggerty_control_2023}, which uses a specially designed static Koopman operator as a pregain term for optimal control. Taking advantage of such techniques will likely have a strong impact towards addressing this issue.

% Another significant limitation of Koopman models is their hefty data requirements. 

\section{Conclusion} \label{sec:conclusion}

In this paper, we have employed a data-driven method to control the shape of tendon-driven continuum soft robots using control affine Koopman models and linear MPC. We show that this method is able to successfully control multi-segment tendon-actuated soft robotic arms by using a per-segment projection, robustly handling the highly nonlinear dynamics that arise from the system.

While these results are promising, they hint at many areas which may require further work. Principled choices of dictionary functions remain an open problem that varies from system to system. Further, the large steady-state error of the Koopman model and high data requirements of Koopman models are problems that may limit the practical application of this class of methods to physical prototypes. As such, in future work, we would like to experimentally validate this approach on physical prototypes, using models learned from characterized and simulated data as in this work, as well as models learned from observed trajectories of the physical prototype itself.

% \section*{Acknowledgments}
% Omitted due to anonymous review

%% Use plainnat to work nicely with natbib. 

\bibliographystyle{mplainnat}
\bibliography{refs}

\end{document}